\documentclass[conference]{IEEEtran}
\IEEEoverridecommandlockouts
\def\BibTeX{{\rm B\kern-.05em{\sc i\kern-.025em b}\kern-.08em
    T\kern-.1667em\lower.7ex\hbox{E}\kern-.125emX}}

\usepackage{times}
\usepackage{xr}

\usepackage[T1]{fontenc}
\usepackage[utf8]{inputenc}
\usepackage{mathtools}

\usepackage{amssymb,mathrsfs}% Typical maths resource packages
\usepackage{amsthm}
\usepackage{bm}
\usepackage{scalerel}
\usepackage{nicefrac}
\usepackage{microtype} 
\usepackage[shortlabels]{enumitem}
\usepackage{graphicx}
\usepackage{epstopdf}
\DeclareGraphicsExtensions{.eps,.png,.jpg,.pdf}

\usepackage{url}
\usepackage{colortbl}
\usepackage{booktabs}
\usepackage{multirow}
\usepackage{colortbl,xcolor}
\usepackage{soul}
\usepackage{xparse,xstring}
\usepackage{calc}
\usepackage{etoolbox}

\makeatletter
\@ifpackageloaded{natbib}{
	\relax
}{
	\usepackage{cite}
}
\makeatother

\usepackage{array}
\newcolumntype{L}[1]{>{\raggedright\let\newline\\\arraybackslash\hspace{0pt}}m{#1}}
\newcolumntype{C}[1]{>{\centering\let\newline\\\arraybackslash\hspace{0pt}}m{#1}}
\newcolumntype{R}[1]{>{\raggedleft\let\newline\\\arraybackslash\hspace{0pt}}m{#1}}

\makeatletter
\let\MYcaption\@makecaption
\makeatother
\usepackage[font=footnotesize]{subcaption}
\makeatletter
\let\@makecaption\MYcaption
\makeatother

\usepackage{glossaries}
\makeatletter
\sfcode`\.1006

\let\oldgls\gls
\let\oldglspl\glspl

\newcommand\fussy@ifnextchar[3]{%
	\let\reserved@d=#1%
	\def\reserved@a{#2}%
	\def\reserved@b{#3}%
	\futurelet\@let@token\fussy@ifnch}
\def\fussy@ifnch{%
	\ifx\@let@token\reserved@d
		\let\reserved@c\reserved@a
	\else
		\let\reserved@c\reserved@b
	\fi
	\reserved@c}

\renewcommand{\gls}[1]{%
\oldgls{#1}\fussy@ifnextchar.{\@checkperiod}{\@}}
\renewcommand{\glspl}[1]{%
\oldglspl{#1}\fussy@ifnextchar.{\@checkperiod}{\@}}

\newcommand{\@checkperiod}[1]{%
	\ifnum\sfcode`\.=\spacefactor\else#1\fi
}

\robustify{\gls}
\robustify{\glspl}
\makeatother

\newacronym{wrt}{w.r.t.}{with respect to}
\newacronym{RHS}{R.H.S.}{right-hand side}
\newacronym{LHS}{L.H.S.}{left-hand side}
\newacronym{iid}{i.i.d.}{independent and identically distributed}
\newacronym{SOTA}{SOTA}{state-of-the-art}
\usepackage{float}

\ifx\notloadhyperref\undefined
	\ifx\loadbibentry\undefined
		\usepackage[hidelinks,hypertexnames=false]{hyperref}
	\else
		\usepackage{bibentry}
		\makeatletter\let\saved@bibitem\@bibitem\makeatother
		\usepackage[hidelinks,hypertexnames=false]{hyperref}
		\makeatletter\let\@bibitem\saved@bibitem\makeatother
	\fi
\else
	\ifx\loadbibentry\undefined
		\relax
	\else
		\usepackage{bibentry}
	\fi
\fi

\usepackage{cleveref-forward}

\crefname{equation}{}{}
\Crefname{equation}{}{}
\crefname{claim}{claim}{claims}
\crefname{step}{step}{steps}
\crefname{line}{line}{lines}
\crefname{condition}{condition}{conditions}
\crefname{dmath}{}{}
\crefname{dseries}{}{}
\crefname{dgroup}{}{}
\crefname{page}{page}{pages}

\crefname{Problem}{Problem}{Problems}
\crefformat{Problem}{Problem~#2#1#3}
\crefrangeformat{Problem}{Problems~#3#1#4 to~#5#2#6}

\crefname{Theorem}{Theorem}{Theorems}
\crefname{Corollary}{Corollary}{Corollaries}
\crefname{Proposition}{Proposition}{Propositions}
\crefname{Lemma}{Lemma}{Lemmas}
\crefname{Definition}{Definition}{Definitions}
\crefname{Example}{Example}{Examples}
\crefname{Assumption}{Assumption}{Assumptions}
\crefname{Remark}{Remark}{Remarks}
\crefname{Rem}{Remark}{Remarks}
\crefname{remarks}{Remarks}{Remarks}
\crefname{Appendix}{Appendix}{Appendices}
\crefname{Supplement}{Supplement}{Supplements}
\crefname{Exercise}{Exercise}{Exercises}
\crefname{TheoremA}{Theorem}{Theorems}
\crefname{CorollaryA}{Corollary}{Corollaries}
\crefname{PropositionA}{Proposition}{Propositions}
\crefname{LemmaA}{Lemma}{Lemmas}
\crefname{DefinitionA}{Definition}{Definitions}
\crefname{ExampleA}{Example}{Examples}
\crefname{RemarkA}{Remark}{Remarks}
\crefname{AssumptionA}{Assumption}{Assumptions}
\crefname{ExerciseA}{Exercise}{Exercises}
\crefname{algorithm}{Algorithm}{Algorithms}
\crefname{figure}{Fig.}{Figs.}
\crefname{table}{Table}{Tables}
\crefname{section}{Section}{Sections}
\crefname{subsection}{Section}{Sections}
\crefname{subsubsection}{Section}{Sections}

\usepackage{algorithm}
\usepackage{algpseudocode}

\ifx\loadbreqn\undefined
	\relax
\else
	\usepackage{breqn}
\fi

\makeatletter
\def\cleartheorem#1{%
    \expandafter\let\csname#1\endcsname\relax
    \expandafter\let\csname c@#1\endcsname\relax
}
\def\clearthms#1{ \@for\tname:=#1\do{\cleartheorem\tname} }
\makeatother

\ifx\renewtheorem\undefined
	\ifx\useTheoremCounter\undefined
		\newtheorem{Theorem}{Theorem}
		\newtheorem{Corollary}{Corollary}
		\newtheorem{Proposition}{Proposition}
		
	\else

	\fi

\fi

\theoremstyle{remark}

\theoremstyle{plain}

\newcommand{\qednew}{\nobreak \ifvmode \relax \else
		\ifdim\lastskip<1.5em \hskip-\lastskip
			\hskip1.5em plus0em minus0.5em \fi \nobreak
		\vrule height0.75em width0.5em depth0.25em\fi}



\newcommand{\nn}{\nonumber\\ }

\NewDocumentCommand{\movedownsub}{e{^_}}{%
	\IfNoValueTF{#1}{%
		\IfNoValueF{#2}{^{}}% neither ^ nor _, do nothing; if no ^ but _, add ^{}
	}{%
		^{#1}% add superscript if present
	}%
	\IfNoValueF{#2}{_{#2}}% add subscript if present
}

\let\latexchi\chi
\RenewDocumentCommand{\chi}{}{\latexchi\movedownsub}

\newcommand{\bF}{\mathbf{F}}

\newcommand{\bI}{\mathbf{I}}

\newcommand{\bL}{\mathbf{L}}

\newcommand{\bM}{\mathbf{M}}

\newcommand{\bp}{\mathbf{p}}

\newcommand{\bW}{\mathbf{W}}
\newcommand{\bx}{\mathbf{x}}
\newcommand{\bX}{\mathbf{X}}

\newcommand{\bY}{\mathbf{Y}}
\newcommand{\bz}{\mathbf{z}}
\newcommand{\bZ}{\mathbf{Z}}

\DeclareSymbolFont{bsfletters}{OT1}{cmss}{bx}{n}
\DeclareSymbolFont{ssfletters}{OT1}{cmss}{m}{n}
\DeclareMathSymbol{\bsfGamma}{0}{bsfletters}{'000}
\DeclareMathSymbol{\ssfGamma}{0}{ssfletters}{'000}
\DeclareMathSymbol{\bsfDelta}{0}{bsfletters}{'001}
\DeclareMathSymbol{\ssfDelta}{0}{ssfletters}{'001}
\DeclareMathSymbol{\bsfTheta}{0}{bsfletters}{'002}
\DeclareMathSymbol{\ssfTheta}{0}{ssfletters}{'002}
\DeclareMathSymbol{\bsfLambda}{0}{bsfletters}{'003}
\DeclareMathSymbol{\ssfLambda}{0}{ssfletters}{'003}
\DeclareMathSymbol{\bsfXi}{0}{bsfletters}{'004}
\DeclareMathSymbol{\ssfXi}{0}{ssfletters}{'004}
\DeclareMathSymbol{\bsfPi}{0}{bsfletters}{'005}
\DeclareMathSymbol{\ssfPi}{0}{ssfletters}{'005}
\DeclareMathSymbol{\bsfSigma}{0}{bsfletters}{'006}
\DeclareMathSymbol{\ssfSigma}{0}{ssfletters}{'006}
\DeclareMathSymbol{\bsfUpsilon}{0}{bsfletters}{'007}
\DeclareMathSymbol{\ssfUpsilon}{0}{ssfletters}{'007}
\DeclareMathSymbol{\bsfPhi}{0}{bsfletters}{'010}
\DeclareMathSymbol{\ssfPhi}{0}{ssfletters}{'010}
\DeclareMathSymbol{\bsfPsi}{0}{bsfletters}{'011}
\DeclareMathSymbol{\ssfPsi}{0}{ssfletters}{'011}
\DeclareMathSymbol{\bsfOmega}{0}{bsfletters}{'012}
\DeclareMathSymbol{\ssfOmega}{0}{ssfletters}{'012}

\makeatletter
\newcommand*\rel@kern[1]{\kern#1\dimexpr\macc@kerna}
\newcommand*\widebar[1]{%
  \begingroup
  \def\mathaccent##1##2{%
    \rel@kern{0.8}%
    \overline{\rel@kern{-0.8}\macc@nucleus\rel@kern{0.2}}%
    \rel@kern{-0.2}%
  }%
  \macc@depth\@ne
  \let\math@bgroup\@empty \let\math@egroup\macc@set@skewchar
  \mathsurround\z@ \frozen@everymath{\mathgroup\macc@group\relax}%
  \macc@set@skewchar\relax
  \let\mathaccentV\macc@nested@a
  \macc@nested@a\relax111{#1}%
  \endgroup
}
\makeatother

\DeclareMathOperator{\var}{var}

\DeclareMathOperator{\cov}{cov}

\newcommand{\ifbcdot}[1]{\ifblank{#1}{\cdot}{#1}}

\DeclarePairedDelimiterX\abs[1]{\lvert}{\rvert}{\ifbcdot{#1}}
\DeclarePairedDelimiterX\parens[1]{(}{)}{\ifbcdot{#1}}
\DeclarePairedDelimiterX\brk[1]{[}{]}{\ifbcdot{#1}}
\DeclarePairedDelimiterX\braces[1]{\{}{\}}{\ifbcdot{#1}}
\DeclarePairedDelimiterX\angles[1]{\langle}{\rangle}{\ifblank{#1}{\cdot,\cdot}{#1}}
\DeclarePairedDelimiterX\ip[2]{\langle}{\rangle}{\ifbcdot{#1},\ifbcdot{#2}}
\DeclarePairedDelimiterX\norm[1]{\lVert}{\rVert}{\ifbcdot{#1}}
\DeclarePairedDelimiterX\ceil[1]{\lceil}{\rceil}{\ifbcdot{#1}}
\DeclarePairedDelimiterX\floor[1]{\lfloor}{\rfloor}{\ifbcdot{#1}}

\DeclareFontFamily{U}{matha}{\hyphenchar\font45}
\DeclareFontShape{U}{matha}{m}{n}{
      <5> <6> <7> <8> <9> <10> gen * matha
      <10.95> matha10 <12> <14.4> <17.28> <20.74> <24.88> matha12
      }{}
\DeclareSymbolFont{matha}{U}{matha}{m}{n}
\DeclareFontSubstitution{U}{matha}{m}{n}

\DeclareFontFamily{U}{mathx}{\hyphenchar\font45}
\DeclareFontShape{U}{mathx}{m}{n}{
      <5> <6> <7> <8> <9> <10>
      <10.95> <12> <14.4> <17.28> <20.74> <24.88>
      mathx10
      }{}
\DeclareSymbolFont{mathx}{U}{mathx}{m}{n}
\DeclareFontSubstitution{U}{mathx}{m}{n}

\DeclareMathDelimiter{\vvvert}{0}{matha}{"7E}{mathx}{"17}
\DeclarePairedDelimiterX\vertiii[1]{\vvvert}{\vvvert}{\ifbcdot{#1}}

\DeclarePairedDelimiterXPP\trace[1]{\operatorname{Tr}}{(}{)}{}{\ifbcdot{#1}} % column vector
\DeclarePairedDelimiterXPP\col[1]{\operatorname{col}}{\{}{\}}{}{\ifbcdot{#1}} % column vector
\DeclarePairedDelimiterXPP\row[1]{\operatorname{row}}{\{}{\}}{}{\ifbcdot{#1}} % row vector
\DeclarePairedDelimiterXPP\erf[1]{\operatorname{erf}}{(}{)}{}{\ifbcdot{#1}}
\DeclarePairedDelimiterXPP\erfc[1]{\operatorname{erfc}}{(}{)}{}{\ifbcdot{#1}}
\DeclarePairedDelimiterXPP\KLD[2]{D}{(}{)}{}{\ifbcdot{#1}\, \delimsize\|\, \ifbcdot{#2}} % KL divergence
\DeclarePairedDelimiterXPP\op[2]{\operatorname{#1}}{(}{)}{}{#2} % general operator

\newcommand{\T}{^{\mathop\intercal}}% transpose notation
\DeclarePairedDelimiterXPP\indicate[1]{{\bf 1}}{\{}{\}}{}{\ifbcdot{#1}}

\NewDocumentCommand\ofrac{s m}{%
	\IfBooleanTF#1%
	{\dfrac{1}{#2}}%
	{\frac{1}{#2}}%
}
\NewDocumentCommand\ddfrac{s m m}{%
	\IfBooleanTF#1%
	{\dfrac{\mathrm{d} {#2}}{\mathrm{d} {#3}}}%
	{\frac{\mathrm{d} {#2}}{\mathrm{d} {#3}}}%
}
\NewDocumentCommand\ppfrac{s m m}{%
	\IfBooleanTF#1%
	{\dfrac{\partial {#2}}{\partial {#3}}}%
	{\frac{\partial {#2}}{\partial {#3}}}%
}

\newcommand{\setgiven}{:}
\providecommand\given{}
\DeclarePairedDelimiterX\Set[2]\{\}{%
	\if#1:%
		\renewcommand\given{\SetSymbol{:}}%
	\else%
		\renewcommand\given{\SetSymbol[\delimsize]{#1}}%
	\fi%
#2
}

\NewDocumentCommand\set{s O{\setgiven} m}{%
	\IfBooleanTF#1%
	{\Set*{#2}{#3}}%
	{\Set{#2}{#3}}%
}

\NewDocumentCommand{\evalat}{ s O{\big} m e{_^} }{%
\IfBooleanTF{#1}%
{\left. #3 \right|}{#3#2|}%
\IfValueT{#4}{_{#4}}%
\IfValueT{#5}{^{#5}}%
}

\providecommand\given{}
\DeclarePairedDelimiterXPP\cprob[1]{}(){}{
\renewcommand\given{\nonscript\,\delimsize\vert\allowbreak\nonscript\,\mathopen{}}%
#1%
}
\DeclarePairedDelimiterXPP\cexp[1]{}[]{}{
\renewcommand\given{\nonscript\,\delimsize\vert\allowbreak\nonscript\,\mathopen{}}%
#1%
}

\DeclareDocumentCommand \P { s e{_^} d() g } {%
	\mathbb{P}%
	\IfBooleanTF{#1}%
		{
			\IfValueT{#2}{_{#2}}%
			\IfValueT{#3}{^{#3}}%
			\IfValueTF{#5}{\cprob{#4 \given #5}}{\IfValueT{#4}{\cprob{#4}}}%
		}%
		{
			\IfValueT{#2}{_{#2}}%
			\IfValueT{#3}{^{#3}}%
			\IfValueTF{#5}{\cprob*{#4 \given #5}}{\IfValueT{#4}{\cprob*{#4}}}%
		}%
}

\DeclareDocumentCommand \E { s e{_^} o g } {%
	\mathbb{E}%
	\IfBooleanTF{#1}%
		{
			\IfValueT{#2}{_{#2}}%
			\IfValueT{#3}{^{#3}}%
			\IfValueTF{#5}{\cexp{#4 \given #5}}{\IfValueT{#4}{\cexp{#4}}}%
		}%
		{
			\IfValueT{#2}{_{#2}}%
			\IfValueT{#3}{^{#3}}%	
			\IfValueTF{#5}{\cexp*{#4 \given #5}}{\IfValueT{#4}{\cexp*{#4}}}%		
		}%
}

\DeclareDocumentCommand \Var { s e{_^} d() g } {%
	\var%
	\IfBooleanTF{#1}%
		{
			\IfValueT{#2}{_{#2}}%
			\IfValueT{#3}{^{#3}}%
			\IfValueTF{#5}{\cprob{#4 \given #5}}{\IfValueT{#4}{\cprob{#4}}}%
		}%
		{
			\IfValueT{#2}{_{#2}}%
			\IfValueT{#3}{^{#3}}%	
			\IfValueTF{#5}{\cprob*{#4 \given #5}}{\IfValueT{#4}{\cprob*{#4}}}%		
		}%
}

\DeclareDocumentCommand \Cov { s e{_^} d() g } {%
	\cov%
	\IfBooleanTF{#1}%
		{
			\IfValueT{#2}{_{#2}}%
			\IfValueT{#3}{^{#3}}%
			\IfValueTF{#5}{\cprob{#4 \given #5}}{\IfValueT{#4}{\cprob{#4}}}%
		}%
		{
			\IfValueT{#2}{_{#2}}%
			\IfValueT{#3}{^{#3}}%	
			\IfValueTF{#5}{\cprob*{#4 \given #5}}{\IfValueT{#4}{\cprob*{#4}}}%		
		}%
}

\ExplSyntaxOn
\NewDocumentCommand \dist {m o o} {%
\mathrm{#1}\left(%
	\IfValueT{#3}{%
		\tl_if_blank:nTF{ #3 }{\cdot\, \middle|\, }{#3\, \middle|\, }%
	}
	\IfValueT{#2}{#2}%
\right)%
}
\ExplSyntaxOff

\NewDocumentCommand {\cbrace} {t+ D[]{black} D(){\widthof{#5}} m m } {%
	\begingroup%
		\color{#2}
		\IfBooleanTF{#1}{%
			\overbrace{#4}^%
		}{
			\underbrace{#4}_%
		}%
		{\parbox[c]{#3}{\centering\footnotesize{#5}}}%
	\endgroup% 
}

\let\oldforall\forall
\renewcommand{\forall}{\oldforall \, }

\let\oldexist\exists
\renewcommand{\exists}{\oldexist \, }

\makeatletter

\newcommand{\rankcolor}[2]{%
	\expandafter\renewcommand\csname #1\endcsname[1]{%
		\ifblank{##1}{%
			{\color{#2} \textbf{#2}}%
		}{%
			\ifmmode
				\textcolor{#2}{\bm{##1}}%
			\else%
				{\color{#2} \textbf{##1}}%
			\fi	
		}%
	}
}

\rankcolor{first}{red}
\rankcolor{second}{blue}
\rankcolor{third}{cyan}
\makeatother

\graphicspath{{./Figures/}{./figures/}}
\DeclareDocumentCommand{\includeCroppedPdf}{ o O{./Figures/} m }{
	\IfFileExists{#2#3-crop.pdf}{}{%
		\immediate\write18{pdfcrop #2#3.pdf #2#3-crop.pdf}}%
	\includegraphics[#1]{#2#3-crop.pdf}
}

\makeatletter
\newcommand*{\addFileDependency}[1]{% argument=file name and extension
  \typeout{(#1)}
  \@addtofilelist{#1}
  \IfFileExists{#1}{}{\typeout{No file #1.}}
}
\makeatother

\definecolor{gray90}{gray}{0.9}
\def\colorlist{red,blue,brown,cyan,darkgray,gray,lightgray,green,lime,magenta,olive,orange,pink,purple,teal,violet,white,yellow}

\makeatletter
\def\startcomment{[}
\ifx\nohighlights\undefined
	\newcommand{\createcolor}[1]{%
			\expandafter\newcommand\csname #1\endcsname[1]{{\color{#1} ##1}}%
	}
	\newcommand{\msout}[1]{\text{\color{green} \st{\ensuremath{#1}}}}
	\newcommand{\del}[1]{{\color{green}\ifmmode \msout{#1}\else\st{#1}\fi}}
\else
	\newcommand{\createcolor}[1]{%
			\expandafter\newcommand\csname #1\endcsname[1]{%
				\noexpandarg%
				\StrChar{##1}{1}[\firstletter]%
				\if\firstletter\startcomment%
					\relax
				\else%
					##1
				\fi
			}%
	}
	\newcommand{\msout}[1]{}
	\newcommand{\del}[1]{}
\fi

\def\@tempa#1,{%
    \ifx\relax#1\relax\else
        \createcolor{#1}%
        \expandafter\@tempa
    \fi
}
\expandafter\@tempa\colorlist,\relax,
\makeatother

\newcommand{\hhide}[1]{}
\ifx\diagnoselabel\undefined
	\relax
\else
	\makeatletter
	\def\@testdef #1#2#3{%
		\def\reserved@a{#3}\expandafter \ifx \csname #1@#2\endcsname
			\reserved@a  \else
			\typeout{^^Jlabel #2 changed:^^J%
				\meaning\reserved@a^^J%
				\expandafter\meaning\csname #1@#2\endcsname^^J}%
			\@tempswatrue \fi}
	\makeatother
\fi

\begin{document}

\title{Modulo Video Recovery via Selective Spatiotemporal Vision Transformer\\

% \thanks{This work was supported by the Singapore Ministry of Education Academic Research Fund Tier 2 grant MOE-T2EP20220-0002.}%
}

\author{%
  \IEEEauthorblockN{Tianyu Geng, Feng Ji, Wee Peng Tay}%
  \IEEEauthorblockA{\textit{Nanyang Technological University}\\
                    Singapore\\
                    \{tianyu.geng, jifeng, wptay\}@ntu.edu.sg}
}

% \title{Conference Paper Title*\\
% {\footnotesize \textsuperscript{*}Note: Sub-titles are not captured in Xplore and
% should not be used}
% \thanks{Identify applicable funding agency here. If none, delete this.}
% }

% \author{\IEEEauthorblockN{1\textsuperscript{st} Given Name Surname}
% \IEEEauthorblockA{\textit{dept. name of organization (of Aff.)} \\
% \textit{name of organization (of Aff.)}\\
% City, Country \\
% email address or ORCID}
% \and
% \IEEEauthorblockN{2\textsuperscript{nd} Given Name Surname}
% \IEEEauthorblockA{\textit{dept. name of organization (of Aff.)} \\
% \textit{name of organization (of Aff.)}\\
% City, Country \\
% email address or ORCID}
% \and
% \IEEEauthorblockN{3\textsuperscript{rd} Given Name Surname}
% \IEEEauthorblockA{\textit{dept. name of organization (of Aff.)} \\
% \textit{name of organization (of Aff.)}\\
% City, Country \\
% email address or ORCID}
% \and
% \IEEEauthorblockN{4\textsuperscript{th} Given Name Surname}
% \IEEEauthorblockA{\textit{dept. name of organization (of Aff.)} \\
% \textit{name of organization (of Aff.)}\\
% City, Country \\
% email address or ORCID}
% \and
% \IEEEauthorblockN{5\textsuperscript{th} Given Name Surname}
% \IEEEauthorblockA{\textit{dept. name of organization (of Aff.)} \\
% \textit{name of organization (of Aff.)}\\
% City, Country \\
% email address or ORCID}
% \and
% \IEEEauthorblockN{6\textsuperscript{th} Given Name Surname}
% \IEEEauthorblockA{\textit{dept. name of organization (of Aff.)} \\
% \textit{name of organization (of Aff.)}\\
% City, Country \\
% email address or ORCID}
% }

\maketitle

\begin{abstract}
Conventional image sensors have limited dynamic range, causing saturation in high-dynamic-range (HDR) scenes. Modulo cameras address this by folding incident irradiance into a bounded range, yet require specialized unwrapping algorithms to reconstruct the underlying signal. Unlike HDR recovery, which extends dynamic range from conventional sampling, modulo recovery restores actual values from folded samples. Despite being introduced over a decade ago, progress in modulo image recovery has been slow, especially in the use of modern deep learning techniques. 
In this work, we demonstrate that standard HDR methods are unsuitable for modulo recovery. Transformers, however, can capture global dependencies and spatial-temporal relationships crucial for resolving folded video frames. Still, adapting existing Transformer architectures for modulo recovery demands novel techniques. To this end, we present Selective Spatiotemporal Vision Transformer (SSViT), the first deep learning framework for modulo video reconstruction. SSViT employs a token selection strategy to improve efficiency and concentrate on the most critical regions. Experiments confirm that SSViT produces high-quality reconstructions from 8-bit folded videos and achieves state-of-the-art performance in modulo video recovery.
\end{abstract}


\section{Introduction}

Conventional image sensors have a limited dynamic range constrained by their well capacity and quantization precision, leading to pixel saturation in scenes with wide light intensity ranges. Capturing such scenes is critical in fields like automotive vision, advanced displays, and photography \cite{hoefflinger2007high,seetzen2004high,reinhard2010high,banterleadvanced}. Since modern sensors' dynamic range is far below natural light intensities \cite{ohta2017smart}, specialized sensors or computational methods are required to address these limitations.

Traditional methods use conventionally captured \emph{\textbf{saturated}} samples to reconstruct images by mapping them to underlying signals \cite{endo2017deep,cao2023unsupervised}. Multi-exposure techniques, such as exposure bracketing \cite{gupta2013fibonacci,hasinoff2016burst} and stacking \cite{wang2019detail}, fuse images at varying exposure levels but struggle with ghosting in dynamic scenes. Multi-sensor systems \cite{tocci2011versatile} mitigate this, but are expensive and complex. Optical encoding \cite{rouf2011glare,metzler2020deep,sun2020learning} and logarithmic irradiance sensors \cite{loose2001self} extend dynamic range at the cost of resolution or precision, while single-shot multi-time imaging sacrifices spatial resolution to achieve similar goals \cite{nayar2003adaptive,hajisharif2015adaptive,serrano2016convolutional,martel2020neural,xu2021deep}.

The aforementioned methods enhance dynamic range but remain constrained. Designing a sensor with an \emph{infinite} dynamic range is impractical due to the finite precision of analog-to-digital converters (ADC). A practical alternative is a \emph{\textbf{modulo}} sensor, which resets its pixel counter to zero upon reaching its maximum capacity, avoiding saturation and allowing continuous photon accumulation.

\begin{figure}[!tb]
  \centering
    \includegraphics[width=0.5\textwidth]{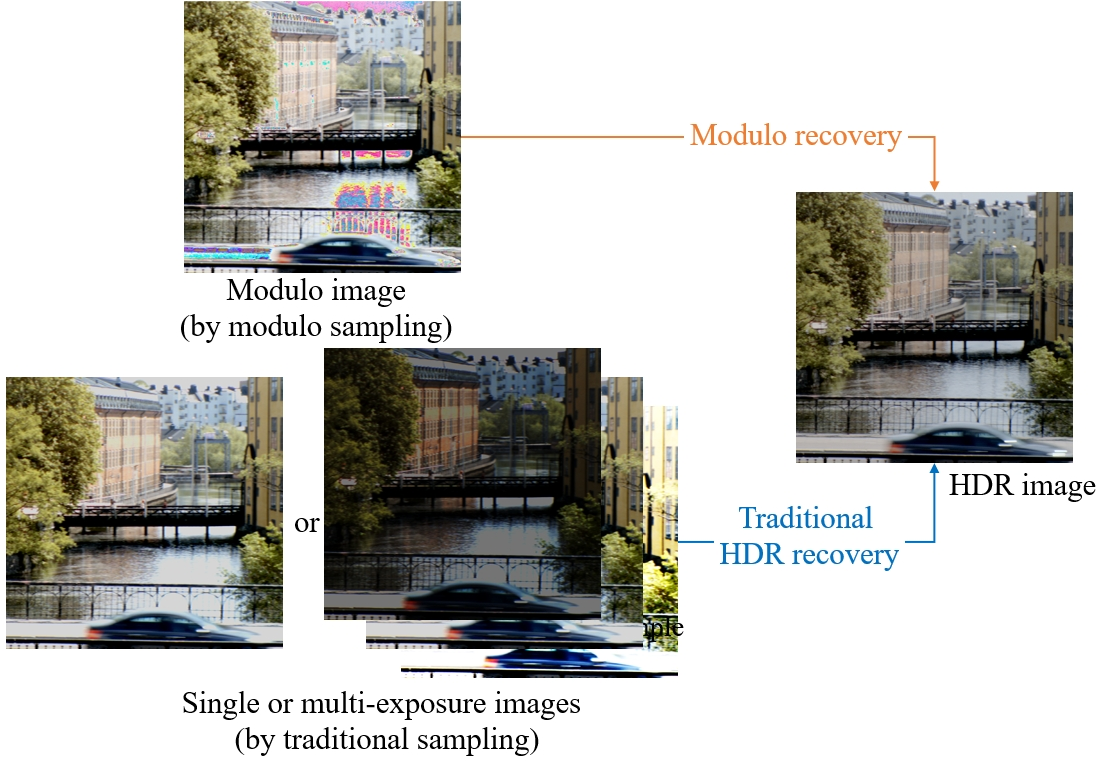}
  \caption{Comparison of two approaches for obtaining HDR images: modulo recovery with modulo-sampled inputs and traditional HDR recovery with single or multi-exposure sampled inputs.}
  \label{fig:teaser}
\end{figure}

In conventional cameras, the analog signal is quantized to $A$ bits by the ADC, with compact cameras typically using 8 bits and high-end DSLRs using 12 or 14 bits. Saturation occurs when the signal exceeds the well capacity. In contrast, a modulo sensor \cite{zhao2015unbounded,zhou2020unmodnet} integrates photocurrent until a threshold is reached, discharges the capacitor, and increments a digital counter. Once the $A$-bit counter reaches its maximum, it resets to zero, providing modulo measurements with theoretically infinite dynamic range. 
\cref{fig:teaser} illustrates the two approaches to obtaining HDR images. Both aim to output HDR images: the upper yellow pathway represents modulo recovery, taking modulo-sampled images as input, while the lower blue pathway represents traditional HDR recovery, typically using single or multi-exposure sampled images as input.

Mathematically, let $\bI$ denote the ground truth image (bit depth $B$), and $\bI_m$ the image recorded by a modulo sensor with $A$ bits ($A < B$). The relationship between them for each pixel at location $[x, y]$ is: 
\begin{align} 
\label{eq:modulo_img} 
& \bI_{m[x,y]} = \bI_{[x,y]} \mod 2^A \quad \text{or } \nn
& \bI_{[x,y]} = \bI_{m[x,y]} + \bL_{[x,y]} \cdot 2^A, \end{align} 
where $\mod$ represents the modulo relation, and $\bL_{[x,y]}$ represents the folding number for pixel $(x, y)$.

\begin{figure}
\centering
\includegraphics[width=0.5\textwidth]{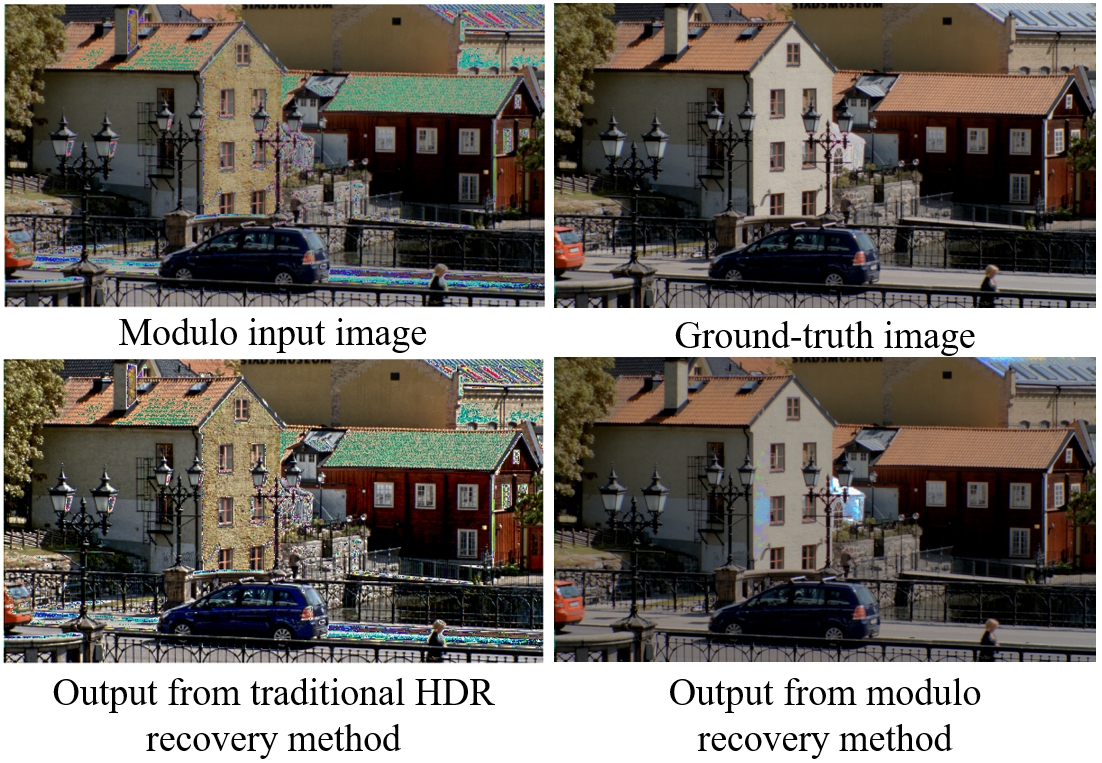}
\caption{Comparison between traditional HDR recovery \protect\cite{tang2023high} and modulo recovery applied to a modulo input. Note that folded pixels (e.g., on the road surface, rooftops, walls, etc.) in the modulo input image are not recovered by the traditional HDR imaging method.}
\label{fig:example}
\end{figure}

Traditional HDR recovery techniques aim to enhance dynamic range but are designed for conventional sampled images, not modulo samples. Traditional methods, such as exposure bracketing \cite{hasinoff2016burst}, stacking \cite{wang2019detail}, or tone mapping \cite{endo2017deep}, rely on combining information from different exposure levels to map low dynamic range (LDR) pixels to HDR values. Some learning-based HDR recovery methods, such as EPCE \cite{tang2023high} and Uformer \cite{wang2022uformer}, explicitly estimate tone mapping functions or leverage Transformer architectures for general image restoration. However, these techniques cannot handle modulo images. \cref{fig:example} compares traditional HDR imaging \cite{tang2023high} and modulo recovery applied to a modulo input. Traditional methods treat folded and unfolded pixels alike, failing to recover the underlying signal. In contrast, modulo recovery identifies folded pixels, estimates folding numbers, and unwraps the signal to achieve accurate reconstruction.

Our work, along with \cite{zhao2015unbounded,zhou2020unmodnet}, focuses on \textbf{restoring images from \emph{modulo} images} captured by \emph{modulo} sensors. We reconstruct the underlying image by identifying modulo pixels and estimating their folding numbers.
Thus, while traditional methods provide only approximate solutions, modulo unwrapping methods theoretically enable unlimited dynamic range imaging. However, reconstruction quality depends heavily on accurate folding number classification.
Existing approaches focused on single-image recovery without leveraging temporal correlations in video reconstruction. To address this limitation, we explore temporal information for modulo video recovery, an increasingly important application in modern contexts.

Recent advances in deep learning, particularly Transformers, have shown exceptional performance in modeling long-range dependencies \cite{arnab2021vivit,zhang2023transferable,metzler2020deep,dosovitskiy2020image,zhang2022improving}, making them promising candidates for tackling challenging tasks like modulo recovery. However, there are two critical questions that need to be addressed in the context of modulo image and video recovery:
\begin{enumerate}[label=\textbf{Q\arabic*}]
    \item \textbf{Can traditional HDR recovery models restore images from modulo images?}
    
    \item \textbf{Can existing Transformer-based architectures be adapted for modulo image recovery?} 
\end{enumerate}
Here, we summarize the answers to these questions: traditional HDR recovery methods and standard video approaches are unsuitable for modulo recovery, and adapting existing Transformer modules with novel strategies is essential to address these challenges. Detailed proofs and comprehensive discussions are presented in the sequel. To answer \textbf{Q1}, in \cref{sec:related_work} and \cref{sec:experiments} we show both analytically and empirically that traditional HDR recovery methods fail to estimate the discrete folding numbers in modulo samples.  
To answer \textbf{Q2}, in \cref{sec:methodology} we demonstrate how standard ViT modules must be adapted—via spatiotemporal attention and our token‐selection strategy—to resolve the ambiguities unique to modulo video reconstruction.

Using well-known Transformer architectures, we propose \textbf{Selective Spatiotemporal Vision Transformer (SSViT)}, a framework that effectively adapts existing Transformer modules to handle the unique properties of modulo image and video recovery. Our key contributions are summarized as follows:
\begin{itemize}
    \item We demonstrate that traditional HDR recovery and standard video approaches are unsuitable for modulo recovery, motivating the development of a tailored framework.
    \item We integrate a Transformer-based architecture into modulo video recovery, enabling simultaneous modeling of spatial and temporal relationships among frames. \emph{An advantage of our approach is the use of existing Transformer modules with novel adaptations, avoiding the need for entirely new architectures.}
    \item We introduce a token selection strategy to improve reconstruction accuracy and memory efficiency, ensuring the practicality of SSViT for real-world applications.
    \item Extensive experiments validate the reliability of SSViT, showing its ability to reconstruct 12-bit videos from 8-bit modulo inputs and outperforming state-of-the-art image recovery methods adapted for modulo tasks.
\end{itemize}

\section{Related Work}\label{sec:related_work}

In this section, we provide an overview of the literature on modulo recovery.

\textbf{Modulo Signal Recovery.  } 
Recovering the original signal from its modulo samples without prior knowledge of the folding number has been a topic of significant research interest. The work \cite{Ayush2017} derived a condition for the recovery of bandlimited signals from folded samples. They also presented a related hardware prototype \cite{Bha21}. The work \cite{JiPraTay:J22} provided conditions for recovering bandlimited graph signals from folded signal samples. The authors in \cite{Gra19} introduced the concept of one-bit unlimited sampling to overcome dynamic range limitations in one-bit quantization, while \cite{Mus18} developed a generalized approximate message passing algorithm for recovering discrete-time sparse signals with noise. The work \cite{Cuc18} studied the denoising of smooth functions observed modulo 1 samples, and \cite{Rudresh2018} addressed folded signal recovery using wavelet transforms, particularly for polynomial signals. All these works do not investigate the use of modern deep learning techniques for modulo signal recovery.

\textbf{Modulo Image Recovery.  }
The unwrapping of modulo images extends 1-D signal recovery techniques to 2-D signals. In optical interferometry, phase unwrapping resolves wrapped phase data, a challenge that also applies to intensity data, requiring specialized algorithms. Robust methods have been developed for phase unwrapping \cite{lang2017robust,shah2017reconstruction,shah2019signal} and natural image recovery \cite{bhandari2020unlimited}. The Markov Random Field (MRF)-based approach \cite{zhao2015unbounded} formulates rollover estimation as an optimization problem, while UnModNet \cite{zhou2020unmodnet} introduces a deep learning framework with a rollover mask predictor using a Convolutional Neural Network (CNN) backbone, which is inefficient for modeling long-range dependencies. However, all these methods focus solely on single-image recovery and lack temporal modeling, limiting their effectiveness for modulo video reconstruction.

\textbf{Modulo Video Recovery.  } To the best of our knowledge, there is no existing work on modulo video recovery. Our work is the first to address this gap.

\section{Methodology} \label{sec:methodology}
In this section, we introduce our proposed SSViT model for modulo video reconstruction from modulo video frames.

\subsection{Motivation}
\label{sec:motivation}
In the task of unwrapping a modulo image, the goal is to predict the number of times each pixel has been folded to recover the underlying HDR image. We define the \emph{class} of a pixel to be its number of folds. Pixels of the same class (i.e., pixels with the same number of folds) can be dispersed throughout the image. Therefore, our neural network needs to be able to identify \emph{long-distance} correlations between pixels, including those in both temporal and spatial dimensions. For this reason, we choose the Transformer architecture as the backbone of our network, as it can learn long-distance correlations between pixels. While some previous computer vision models for videos, such as TDNet \cite{TDNetHu2020temporally}, employ an attention module to capture temporal relations, they are limited to focusing on the nearest one to three frames. This is because collecting too many frames would result in a significantly large token quantity, thereby hindering the ability to capture temporal relations over longer time ranges.

As a pixel-wise classification problem, in addition to the aforementioned challenge, there are three other main challenges:
\begin{itemize}
    \item A large number of classes.
    \item Interspersed pixels of different classes.
    \item Clustering of pixels of the same class in small regions.
\end{itemize}
Moreover, convolutional and pooling operations that result in information loss further exacerbate these challenges.

\begin{figure}
  \centering
    \includegraphics[width=0.25\textwidth]{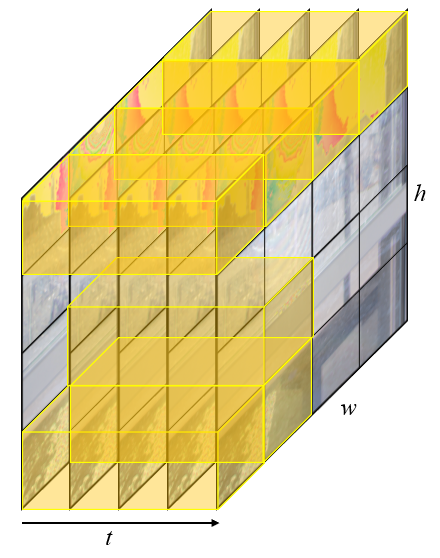}
  \caption{Illustration of token selection. The transparent yellow tubes represent the intricate areas selected by the token selection algorithm, where the folding number varies significantly and undergoes noticeable changes across nearby frames. Note that the segmentation of patches and the number of tokens selected in the illustration are for demonstration purposes only. The actual configurations may vary based on experimental requirements.}
  \label{fig:motivation}
\end{figure}

To tackle challenge 1, considering the specificity of pixel classes in the task (i.e., the number of folds per pixel), we adopt an iterative prediction strategy. However, the use of an iterative Transformer framework leads to a high computational cost, resulting in long training and inference times. To tackle challenges 2 and 3, and reduce the computational cost resulting from the large token quantity and iterative Transformer framework, we adopt an approach that processes intricate tokens more thoroughly while streamlining the processing of basic tokens. To accomplish this, we introduce a token selection strategy, which simultaneously enhances the efficiency and accuracy of our neural network for this task. We improve the network's ability to capture temporal relationships by categorizing the areas in each video clip into ``basic'' and ``intricate''. Intricate areas typically include multiple semantic categories, and capturing their temporal relations can improve precision and temporal stability. However, providing basic areas with temporal information may not have much effect. To identify intricate areas within the video clip, we introduce a method called Neighboring Similarity Matrix (NSM) presented in \cite{li2021video}, which combines cosine distance and Kullback-Leibler (KL) divergence. \cref{fig:motivation} shows an example where intricate tokens are selected.

For each token selected in the target frames, we extract non-overlapping, spatiotemporal ``tubes'' from the input video clip. By integrating the token selection into our spatiotemporal Transformer, we have developed a novel method for modulo video reconstruction. Our approach can capture temporal relations to maintain temporal consistency and balance accuracy and inference efficiency.

\subsection{Problem Formulation}
\label{sec:problem}

Consider an HDR video frame sequence with each frame having pixel dimensions $X \times Y$ and $C$ color channels. We represent the brightness of the $T$-th frame as $\bF_T \in \mathbb{R}^{X\times Y\times C}$ with 
\begin{align}
\bF_T = \{ \bF_{T[x, y, c]} \mid 1 \leq x \leq X, 1 \leq y \leq Y, 1 \leq c \leq C \}
\end{align}
and its corresponding modulo observation as $\bF_{T,m}=\{\bF_{T,m[x,y,c]}\}$, where ${[x,y]}$ are the pixel coordinates and $c$ denotes the color channel index. $\bF_{T,m}$ consists of the least significant $A$ bits (i.e., lower bits typically in a binary encoder) of $\bF_{T}$.  
The relation between $\bF_{T}$ and $\bF_{T,m}$ can be expressed as:
\begin{align} 
\label{eq:modulo}
& \bF_{T,m}= \bF_{T} \mod 2^{A} \quad \text { or } \quad \nn
& \bF_{T}=\bF_{T,m}+2^{A} \cdot \mathbf{L}, \ \ 0\leq \bF_{T,m[x,y,c]} < 2^A,
\end{align}
where $\mathbf{L}=\{\bL_{[x,y,c]}\}$ denotes the folding numbers for all pixels.

Our model takes as input a modulo video clip $\mathcal{F}_m$ comprising $n_{\text{c}}$ modulo frames, and we want to recover the ground truth clip $\mathcal{F}$. Specifically, at each frame index $T$, we aim to recover 
$\mathcal{F}=\{\bF_{T-n_{\text{c}}},\cdots,\bF_{T}\}$ by unwrapping the corresponding modulo clip $\mathcal{F}_m =\{\bF_{T-n_{\text{c}},m},\cdots, \bF_{T,m}\}$ captured by a modulo camera. By Eq. \cref{eq:modulo}, our objective is equivalent to estimating the folding numbers $\mathcal{L}=\{\mathbf{L}_{T-n_{\text{c}}},\cdots,\mathbf{L}_{T}\}$. Additionally, during the reconstruction process, we set the sliding window step size to 1 instead of $n_{\text{c}}+1$ to achieve a smoother visual effect.

Interpreted probabilistically, our goal is to estimate $\text{argmax}_\mathcal{L}~p\left(\mathcal{L} \mid \mathcal{F}_{m}\right)$. Theoretically, if each entry of $\mathcal{L}$ is considered as the label of the corresponding pixel, it belongs to the space of non-negative integers $\mathbb{N}$. 
Estimating the likelihood directly poses a challenge due to the infinite and discrete label space. %{\color{red} [I think even if it is finite, it is still challenging. Maybe you should argue due to ``discreteness''.]}

To enhance model tractability, we employ a factorization approach over the folding numbers $\mathcal{L}$. 
Let $\mathcal{M}^{(k)}=\{\bM^{(k)}_{T-n_{\text{c}}},\cdots,\bM^{(k)}_{T}\}$, where $\bM^{(k)}_{T} \in \{0,1\}^{X\times Y\times C}$ is a binary folding map in the $k$-th order that 
\begin{align}
    \mathcal{M}^{(k)}_{[x,y,c,t]}=\left\{
    \begin{array}{ll}
        1 & \text { if } k \leq \mathcal{L}_{[x,y,c,t]} \\
        0 & \text { otherwise }
    \end{array}%
    \text { and } \sum_{k=1}^{\infty} \mathcal{M}^{(k)}=\mathcal{L}.
\right. 
\end{align}

Then, we have 
\begin{align}
\label{eq:prob_goal}
    &p\left(\mathcal{L} \mid \mathcal{F}_{m}\right) 
    = p\left(\sum_{k=1}^{\infty} \mathcal{M}^{(k)} \mid \mathcal{F}_{m}  \right) \nn
    &= p\left(\mathcal{M}^{(1)} \mid \mathcal{F}_{m}\right) \prod_{k=1}^{\infty} p\left(\mathcal{M}^{(k+1)} \mid \mathcal{M}^{(1)}, \ldots, \mathcal{M}^{(k)}, \mathcal{F}_{m}\right). 
\end{align}%
Referring to Eq. \cref{eq:modulo}, with the prediction of the binary folding masks $\mathcal{M}^{(k)}$, we can update the modulo frames $\mathcal{F}^{(k)}_{m}$ in the $k$-th order of the target modulo frames $\mathcal{F}_{m}$ as 
\begin{align}
\label{eq:reconstruction}
    \mathcal{F}_{m}^{(k)}=\mathcal{F}_{m}+2^{A} \cdot\left(\mathcal{M}^{(1)}+\cdots+\mathcal{M}^{(k)}\right).
\end{align}
We define $\mathcal{F}^{(k)}_{m}=\{\bF^{(k)}_{T-n_{\text{c}},m}, \cdots, \bF^{(k)}_{T,m}\}$, and $\mathcal{F}^{(0)}_{m}=\mathcal{F}_m$ for simplicity, then we have $p\left(\cdot \mid \mathcal{M}^{(1)}, \ldots, \mathcal{M}^{(k)}, \mathcal{F}_{m}\right)=p\left(\cdot \mid \mathcal{F}^{(k)}_{m} \right)$. And
Eq. (\ref{eq:prob_goal}) can be further written as
\begin{align}
    p\left(\mathcal{L} \mid \mathcal{F}_{m}\right)=\prod_{k=0}^{\infty} p\left(\mathcal{M}^{(k+1)} \mid \mathcal{F}_{m}^{(k)}\right).
\end{align}
In the $k$-th order, estimating $p\left(\mathcal{M}^{(k+1)} \mid \mathcal{F}_{m}^{(k)}\right)$  is essentially about estimating the corresponding binary folding mask once we have an updated modulo clip $\mathcal{F}_{m}^{(k)}$. This transforms the original problem into an iterative process of per-pixel binary labeling, which stops when $\mathcal{M}^{(k+1)}=\mathbf{0}$.

The proposed model iteratively updates the modulo frames $\mathcal{F}_{m}$ by predicting the binary folding mask $\mathcal{M}^{\left( k \right)}$ in each $k$-th order, and produces the HDR result $\mathcal{F}$. The algorithm proceeds iteratively until it halts. Each cycle can be represented as follows:
\begin{align}
\label{eq:rec_iter}
    \mathcal{F}_{m}^{(k+1)}=\mathcal{F}_{m}^{(k)}+2^{A} \cdot \mathcal{M}^{(k+1)}=\mathcal{F}_{m}^{(k)}+g\left( \mathcal{F}_{m}^{(k)} \right),
\end{align}
where $g$ represents the prediction model.

\subsection{SSViT}
\label{sec:method}

\begin{figure*}
  \centering
    \includegraphics[width=\textwidth]{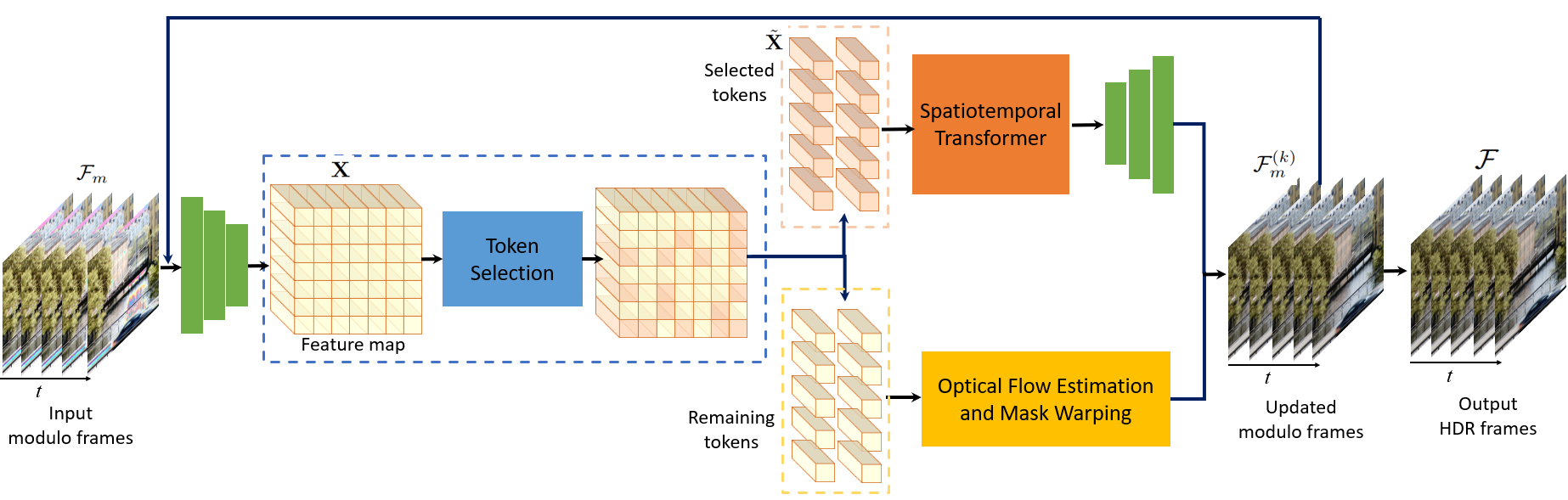}
  \caption{The flowchart of the proposed Selective Spatiotemporal Vision Transformer. Initially, an encoder is employed to convert all frames of the input video clip into embedding features. These embedding features serve as the basis for determining selected areas in the video clip using our token selection strategy. Following this, the spatiotemporal Transformer is utilized to grasp temporal relationships, while unselected tokens undergo optical flow estimation and mask warping operations. Ultimately, the output from the spatiotemporal Transformer is directed to a decoder to predict the folding mask.}
  \label{fig:structure}
\end{figure*}

Our model's overall flowchart is depicted in \cref{fig:structure}. The model takes as input a video clip comprising $T$ modulo frames $\mathcal{F}_m=\{\bF_{T-n_{\text{c}},m}, \cdots, \bF_{T,m}\}$, where the frames $\mathcal{F}_m$ are used by our model in leveraging temporal relations, and $\mathcal{F}$ will be predicted parallel. During the inference phase, frames are encoded only once, and the high-level features are reused by the spatiotemporal Transformer, minimizing redundant computations and ensuring parallel reconstruction.

Our network comprises four main components: the shared encoder, the token selection module, the Transformer-based backbone, and the folding number prediction decoder. The shared encoder, the token selection module, and the mask prediction decoder are fundamental parts, designed to extract the information from each frame and estimate folding numbers per pixel, respectively.

As we iteratively predict the folding number using a recursive approach, prediction errors may accumulate during the prediction process. To address this issue, we design an efficient token selection strategy for the target frames $\mathcal{F}_m$ to focus our neural network on more valuable areas. Given the relatively high computational complexity of the Vision Transformer networks,  the token selection of the spatiotemporal Transformer also reduces the overall computational complexity of the spatiotemporal Transformer without compromising the performance of the mask prediction model. After the token selection process, the unselected tokens are assigned folding number masks by leveraging an optical flow approach using historical frames. Specifically, we employ FlowNet2 \cite{ilg2017flownet} to estimate the optical flow map, which allows us to accurately track the motion between frames. By applying this estimated flow map, we perform a warping operation to retrieve the necessary folding number mask from the corresponding historical frame.

The Transformer architecture comprises Layer Normalization (LN), multi-head self-attention (MSA), and Multi-layer Perceptron (MLP) blocks. Embedding features are divided into non-overlapping windows, with MSA executed within each window.

While standard Transformers are effective for conventional image recovery, directly applying them to modulo images without modification is infeasible due to the unique challenges of folded data. The inherent ambiguities in modulo images, caused by periodic folding of irradiance values, make it difficult for a vanilla Transformer to distinguish between different folding states. To address this, we introduce a token selection strategy that filters out redundant information and focuses computational resources on intricate regions, improving the model’s ability to recover the correct underlying signal. Additionally, our framework adapts a spatiotemporal Transformer by removing positional embeddings, to exploit temporal correlations across video frames, ensuring more accurate reconstruction.

This design effectively adapts Transformer-based architectures for modulo recovery, answering \textbf{Q2} and demonstrating that existing Transformer modules, when properly modified, can successfully handle the unique challenges of modulo image and video restoration. In the following section, we provide details of the token selection and the spatiotemporal Transformer.

\subsubsection{Token Selection}
\label{sec:tokenselection}

Inspired by \cite{marin2019efficient}, we propose a token selection mechanism to refine predictions efficiently. 
The embedding feature $\bX \in \mathbb{R}^{l \times h \times w \times d_q}$ of the current frame is partitioned into basic and intricate areas. 
Basic areas contain consistent folding labels, while intricate areas exhibit variability. 
We focus computational resources on intricate areas using an extended 3D Neighboring Similarity Matrix (NSM) based on cosine distance and KL divergence.

For a neighborhood radius $r$, at coordinate $[s, u, v]$, the feature vector $\bx \in \mathbb{R}^{1 \times d_q}$ is extracted, 
along with a local neighborhood $\bX_\text{neighbor}$, flattened into $\bX_\text{local} \in \mathbb{R}^{n_b \times d_q}$, where $n_b = (2r+1)^3$. 
The similarity distribution $\bp_\text{sim}$ is computed via:
\begin{align}
    \bp_{\text{sim}} = \mathrm{SoftMax}\left(\bX_\text{local} \cdot \bx\T\right).
\end{align}

The NSM is defined as:
\begin{align}
    \mathcal{D}_\text{NSM} = \mathcal{D}_\text{KL} + \mathcal{D}_\text{cos},
\end{align}
where $\mathcal{D}_\text{KL}$ evaluates the divergence between $\bp_\text{sim}$ and a uniform distribution $\bp_\text{u}$:
\begin{align}
    \mathcal{D}_\text{KL} = \mathrm{KL}\left(\bp_\text{u} \| \bp_\text{sim}\right) = \sum_{i=1}^{n_b} \bp_{\text{u}[i]} \log \frac{\bp_{\text{sim}[i]}}{\bp_{\text{u}[i]}},
\end{align}
and $\mathcal{D}_\text{cos}$ measures cosine similarity:
\begin{align}
    \mathcal{D}_\text{cos} = \frac{1}{n_b} \sum_{i=1}^{n_b}\left(1 - \frac{\bX_{\text{local}[i,:]} \cdot \bx\T}{\|\bX_{\text{local}[i,:]}\|_2 \|\bx\|_2}\right).
\end{align}

High $\mathcal{D}_\text{NSM}$ values indicate intricate areas requiring attention.  
The KL divergence term quantifies how much the local similarity scores deviate from uniformity, which highlights regions where folding patterns vary sharply (e.g., fold edges). The cosine-distance term measures the average angular dissimilarity of feature vectors, capturing abrupt directional changes in the embedding space. The combined NSM thus robustly isolates intricate folding regions from homogeneous patches.

Averaging $\mathcal{D}_\text{NSM}$ across dimensions yields $\bar{\mathcal{D}}_\text{NSM}$, which is used to extract tokens with higher complexity. 
This results in the selected token set $\tilde{\bX} \in \mathbb{R}^{\tilde{n} \times d_q}$.

\subsubsection{Spatiotemporal Transformer}

To capture high-level spatiotemporal information from the selected tokens, we employ a Transformer encoder inspired by ViT \cite{dosovitskiy2020image} with joint space-time attention \cite{arnab2021vivit}. Our token selection strategy ensures computational efficiency by reducing input size while retaining essential information. The encoder processes $n$ spatiotemporal ``tubes'' from the video input, rasterizing them into one-dimensional tokens $\bz_i \in \mathbb{R}^d$, which are combined into a sequence:
\begin{align}
    \bZ = [\bW\bX_1,\ \bW\bX_2,\dots,\ \bW\bX_n],
\end{align}
where $\bW$ functions similarly to a two-dimensional convolution operator. 

Unlike the standard ViT, we remove positional embedding, as token positions are not critical for our task. Each Transformer layer $\ell$ consists of Multi-Headed Self-Attention (MSA), Layer Normalization (LN), and Multi-Layer Perceptron (MLP) blocks, defined as:
\begin{align}
    \bY^{\ell} = \mathrm{MSA}\left(\mathrm{LN}\left(\bZ^{\ell}\right)\right)+\bZ^{\ell},
\end{align}
\begin{align}
    \bZ^{\ell+1} = \mathrm{MLP}\left(\mathrm{LN}\left(\bY^{\ell}\right)\right)+\bY^{\ell}.
\end{align}

The MLP includes two linear projections separated by a GELU activation \cite{hendrycks2016gaussian}, maintaining a constant token dimension across layers.

\subsection{Training and Inference}

During the training stage, the cross-entropy loss function is utilized to train the proposed SSViT. During the inference phase, SSViT iteratively predicts the binary folding mask $\mathcal{M}^{\left( k \right)}$ required for modulo video reconstruction (Eq. (\ref{eq:reconstruction})), ultimately completing the modulo video reconstruction. The algorithmic procedure is depicted in Algorithm \ref{alg:inference}.

\begin{algorithm}[H]
\caption{Inference}
\label{alg:inference}
\begin{algorithmic}
\State{\textbf{Input:} A trained SSViT denoted as $g$, a modulo video clip $\mathcal{F}_m$ with $A$ bits, target higher bit depth $B$, iteration count $k=1$}
\end{algorithmic}

\begin{algorithmic}[1]
\While{$k <  2^{B-A}$}
\State{Predict the binary folding mask $\mathcal{M}^{\left( k \right)}$ using $g$}
\State{Calculate $\mathcal{F}_{m}^{(k+1)}$, which is the input of SSViT for the next iteration using Eq. (\ref{eq:rec_iter})}
\State{$k= k+1$}
\EndWhile
\State{$\hat{\mathcal{F}} = \mathcal{F}_{m}^{(k+1)}$}
\end{algorithmic}

\begin{algorithmic}
\State{\textbf{Output:} The reconstructed clip $\hat{\mathcal{F}}$ with $B$ bits}
\end{algorithmic}
\end{algorithm}

\section{Experiments} \label{sec:experiments}

\subsection{Experimental Setup}
\label{sec:setup}
We employ PyTorch to implement our proposed approach. The experiments are conducted for 200,000 iterations, utilizing hardware comprising a 12th Gen Intel Core i7-12700KF processor and 1 NVIDIA GeForce RTX 4090 GPU. The clip length $n_{\text{c}}$ is set to 4 empirically. We utilize the Adam optimizer with an initial learning rate set to 1e-4. For tokens not selected by the token selection strategy, we use FlowNet2 \cite{ilg2017flownet} to predict their folding number mask from historical frames. All HDR images appearing in this article have been tone-mapped to facilitate presentation in the paper. For fair comparison, we maintain consistent training and testing conditions across different methods.

\textbf{Tone Mapping.}  
Most tone mapping operators are designed for static images, scaling HDR luminance independently for each frame in video sequences. This often causes abrupt changes in brightness between frames, resulting in noticeable flickering. To address this, we compute the average brightness of each frame and adjust it based on the previous frame to maintain temporal consistency. Additionally, we apply Reinhard tone mapping using the global average brightness of the entire video to ensure uniform adjustment across the sequence, effectively reducing flickering artifacts.

\textbf{Datasets.}  
Learning-based methods, particularly Transformers, require substantial training data. Since no public dataset exists for our task, we collected HDR images from various video sources \cite{kalantari2019deep,froehlich2014creating,perez2021ntire} for training and used the LiU \cite{kalantari2019deep} and HdM \cite{froehlich2014creating} datasets for testing.

The LiU dataset includes 12 HDR video sequences and 2 light probe sequences, downscaled to $1280 \times 720$ resolution. The HdM dataset provides diverse scenes with a dynamic range of up to 18 stops at $1920 \times 1080$ resolution. Both datasets store frames in OpenEXR format, offering varied environments and brightness changes.

% The NTIRE 2021 HDR dataset \cite{perez2021ntire} is a benchmark dataset designed for the NTIRE 2021 HDR Challenge, which is a competition that focuses on enhancing the quality of HDR images. The resolution of the HDR video is $1900\times 1060$. All frames are stored in .png file format.

\textbf{Data Preparation.}  
Building on the single-image pipeline in \cite{zhou2020unmodnet}, we extend data preparation to video. For each frame, we generate a modulo sample, a binary folding mask, updated modulo frames, and an LDR frame. Key steps include re-exposing HDR frames to regulate over-exposure rates, quantizing to $B$ bits for ground truth, and deriving corresponding $A$-bit modulo frames, folding numbers, and binary masks:
\begin{align}
    \bF_{T,m} = \bF_{T} \mod 2^A, \quad \mathbf{L} = \frac{\bF_{T} - \bF_{T,m}}{2^A}.
\end{align}
The updated modulo frames and binary folding masks are then computed iteratively. Our approach ensures consistent exposure and prepares data tailored for video tasks.

% \textbf{Pre-training.}  
% We use the pre-trained VideoMAE model from \cite{tong2022videomae}, which employs a vanilla ViT backbone with joint space-time attention and tube embedding. Pre-trained on a masked video reconstruction task with a $90\%$ masking ratio, this model does not rely on additional large-scale datasets.

\subsection{Evaluation}
\label{sec:evaluation}

We evaluate the reconstructed images using two metrics: Peak Signal-to-Noise Ratio (PSNR) and Structural SIMilarity (SSIM), following \cite{zhou2020unmodnet}. Pixels near the image boundary are excluded during testing. For a ground-truth frame $\bF \in \mathbb{R}^{a \times b}$ and an estimated frame $\widehat{\bF} \in \mathbb{R}^{a \times b}$, PSNR and SSIM are defined as:
\begin{align} 
\text{PSNR}(\bF,\widehat{\bF})=20\log_{10}\left( \frac{\max(\bF)}{\sqrt{\text{MSE}(\bF,\widehat{\bF})}} \right), 
\end{align}
\begin{align} 
\text{SSIM}(\bF,\widehat{\bF})=\frac{(2\mu_{\bF}\mu_{\widehat{\bF}}+c_1)(2\sigma_{\bF\widehat{\bF}}+c_2)}{(\mu_{\bF}^2+\mu_{\widehat{\bF}}^2+c_1)(\sigma_{\bF}^2+\sigma_{\widehat{\bF}}^2+c_2)},
\end{align}
where $\mu$, $\sigma$, and $\sigma_{\bF\widehat{\bF}}$ represent the mean, variance, and covariance, and $c_1$, $c_2$ are constants to avoid division instability.

%The FSIM \cite{FSIM2011fsim} is the Fourier style of SSIM, which extracts feature points of human interest based on phase congruency and image gradient magnitude to evaluate image quality. The VIF \cite{VIF2006image} treats the human visual system as a communication channel and predicts the subjective image quality by computing the mutual information between the reconstructed image and the reference image. 

\begin{figure*}
\centering
\includegraphics[width=\textwidth]{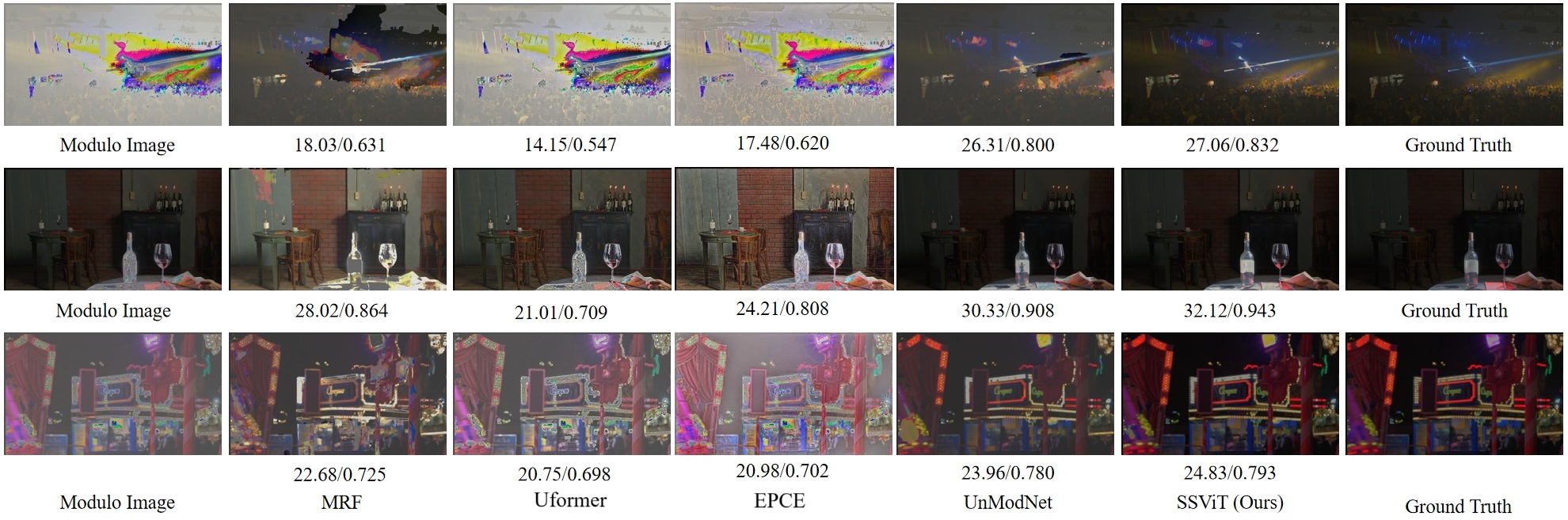}
\caption{Results on the HdM dataset. Quantitative evaluations using PSNR (dB) / SSIM are displayed below each image. }
\label{fig:comparison2}
\end{figure*}

\begin{table}[!htb]
\centering                       
\setlength{\tabcolsep}{4pt}           
\renewcommand{\arraystretch}{1.3}     
\caption{Quantitative results on synthetic data: our SSViT vs.\ Uformer, EPCE, MRF and UnModNet.}\label{tab:syn}

\begin{tabular}{lcccccc}
\toprule
             Dataset             &   Metric   & Uformer  & EPCE   & MRF   & UnModNet & SSViT  \\ 
\midrule
\multirow{2}{*}{LiU} & PSNR(dB) & 11.38  & 13.59 & 12.43 & 27.71    & \textbf{28.85} \\ 
                          & SSIM    & 0.482  & 0.597 & 0.493 & 0.811    & \textbf{0.871} \\ 
\midrule
\multirow{2}{*}{HdM} & PSNR(dB)  & 14.36  & 16.03     & 15.84 & 28.03    & \textbf{29.38} \\ 
                          & SSIM  &  0.535 & 0.589    & 0.577 & 0.824    & \textbf{0.850} \\ 
\bottomrule
\end{tabular}
\end{table}

\cref{fig:comparison2} present a comparative analysis of SSViT against various reconstruction algorithms, assessing both qualitative and quantitative aspects. Our findings, as depicted in \cref{fig:comparison2} and \cref{tab:syn}, demonstrate the superior performance of our model over Uformer \cite{wang2022uformer}, EPCE \cite{tang2023high}, the MRF-based algorithm \cite{zhao2015unbounded}, and UnModNet \cite{zhou2020unmodnet}, which is currently the state-of-the-art modulo image recovery method, in capturing richer details.

% \begin{figure*}
%   \centering
%     \includegraphics[width=\textwidth]{Figures/Comparison_LiU.png}
%   \caption{Results on the LiU dataset. Quantitative evaluations using PSNR (dB) / SSIM are displayed below each image. We refer the reader to \cite{geng23366272} for high-resolution image examples. Differences between the outputs from different methods are also highlighted in the Github repository. }
%   \label{fig:comparison}
% \end{figure*}

It is worth noting that Uformer, EPCE, the MRF-based method, and UnModNet are not specifically designed for videos; unlike SSViT, they process individual frames rather than entire video clips. Additionally, Uformer and EPCE differs from the other three methods as it belongs to the traditional HDR reconstruction category, utilizing LDR frames as input rather than modulo frames. 
In contrast, modulo frames sample saturated pixels as modulo pixel values, whereas LDR frames sample saturated pixels as maximum pixel values. The visual distinction of the inputs can be observed in \cref{fig:teaser}. 

The challenge of conventional HDR reconstruction lies in estimating and predicting missing brightness and color information based on the limited information provided by LDR frames. On the other hand, modulo video reconstruction faces challenges similar to pixel-wise classification, mainly in determining whether pixels are folded and how many times they have been folded. Therefore, our comparison aims to provide insight into the differences between modulo video reconstruction and conventional HDR reconstruction, although it may not be entirely equitable. This also answers the \textbf{Q1} raised in the introduction: HDR recovery techniques, designed for conventional inputs, often assume continuous frame dependencies, failing to estimate the discrete folding numbers in modulo frames, namely that traditional HDR recovery methods are ineffective for the modulo recovery task.

Based on the visual experiment results, all algorithms encounter difficulties in accurately recovering modulo pixels when they are disordered or form abstract patterns in the modulo frame. This observation suggests an inherent limitation of such methods.

Furthermore, it is evident that methods like EPCE, representing HDR reconstruction from LDR videos, although not encountering incorrectly recovered modulo pixels like methods relying on modulo videos, still face challenges due to the loss of brightness information in the input LDR video caused by pixel saturation. This leads to suboptimal overall video recovery, particularly in areas of extreme brightness, as indicated by lower PSNR and SSIM scores compared to SSViT.

% Due to space limitations, we present only a subset of the experimental results. We refer the reader to the supplementary material for additional results. Furthermore, video comparisons can be viewed at \href{https://drive.google.com/drive/folders/1Dn4G8Fqm6DMJEdju3p5F2J8XBhCP7d5w?usp=sharing}{this link}.

% Due to space limitations, we provide additional experimental results and technical details at \href{https://drive.google.com/drive/folders/1Dn4G8Fqm6DMJEdju3p5F2J8XBhCP7d5w?usp=sharing}{this link}, where readers can find extended analyses and video comparisons.

Due to space limitations, we provide additional experimental results and technical details at the following link:  
\url{https://sites.google.com/view/supplementaryijcnnid5522}, where readers can find extended analyses and video comparisons.  

% and \url{https://github.com/geng23366272/folded-HDR-video-reconstruction}

It can be observed that SSViT achieves superior reconstruction performance. For example, in \textit{Bridge.mp4}, SSViT produces more accurate reconstructions of both the water and the road surface compared to other methods, demonstrating its effectiveness in leveraging the long-range dependencies in modulo inputs to better estimate the folding numbers and preserving structural consistency, thereby enabling precise HDR reconstruction.

\section{Conclusion} \label{sec:conclution}

To recover videos from modulo observations, this work proposes a Transformer-based deep learning approach tailored for video data. To the best of our knowledge, it is the first to employ a Transformer model for modulo video reconstruction. Specifically, a Selective Spatiotemporal Transformer network is introduced to capture spatiotemporal dependencies across frames, augmented by a token selection strategy that boosts reconstruction performance and memory efficiency. Experimental findings confirm that the proposed method achieves state-of-the-art results on modulo video reconstruction.

Despite its effectiveness, our approach relies on pre-training and may struggle with extreme folding cases. Future work includes improving robustness to highly folded regions and extending the framework to real-world modulo cameras.

%% The file named.bst is a style file for BibTeX 0.99c

\bibliographystyle{IEEEtran}
\bibliography{bib/References}

\end{document}